\documentclass[11pt,letterpaper]{article}
\usepackage{ijcnlp2017}
\usepackage{times}
\usepackage{latexsym}
\usepackage{graphicx}
\usepackage{booktabs}
\usepackage{todonotes}
\ijcnlpfinalcopy

\title{\textsc{All-In-1}: Short Text Classification with One Model for All Languages}

\author{Barbara Plank \\
Center for Language and Cognition\\
University of Groningen\\
  {\tt b.plank@rug.nl} }

\date{}

\begin{document}
\maketitle
\begin{abstract}
We present \textsc{All-In-1}, a simple model for multilingual text classification that does not require any parallel data. It is based on a traditional Support Vector Machine classifier exploiting multilingual word embeddings and character n-grams. Our model is simple, easily extendable yet very effective, overall ranking 1st (out of 12 teams) in the IJCNLP 2017 shared task on customer feedback analysis in four languages:\ English, French, Japanese and Spanish. 
  \end{abstract}

\section{Introduction} 

Customer feedback analysis is the task of classifying short text messages into a set of predefined labels (e.g., bug, request). It is an important step towards effective customer support. 

However, a real bottleneck for successful classification of customer feedback  in a multilingual environment is the limited transferability of such models, i.e., typically each time a new language is encountered a new model is built from scratch. This is clearly impractical, as maintaining separate models is cumbersome, besides the fact that existing annotations are simply not leveraged. 

In this paper we present our submission to the IJCNLP 2017 shared task on customer feedback analysis, in which data from four languages was available (English, French, Japanese and Spanish).
Our goal was to build a single system for all four languages, and compare it to the traditional approach of creating separate systems for each language. We hypothesize that a single system is beneficial, as it can provide positive transfer, particularly for the languages for which less data is available.
The contributions of this paper are:
\begin{itemize}
\setlength\itemsep{-1pt}
\item We propose a very simple multilingual model for four languages that overall ranks first (out of 12 teams) in the IJCNLP 2017 shared task on Customer Feedback Analysis.
\item We show that a traditional model outperforms neural approaches in this low-data scenario. \item We show the effectiveness of a very simple approach to induce multilingual embeddings that does not require any parallel data. 
\item Our \textsc{All-In-1} model is particularly effective on languages for which little data is available.
\item Finally, we compare our approach to automatic translation, showing that translation negatively impacts classification accuracy.
\item To support reproducibility and follow-up work all code is available at: {\url{https://github.com/bplank/ijcnlp2017-customer-feedback}}
\end{itemize}

\section{\textsc{All-In-1}: One Model for All}

\begin{figure}[h!]\center
\includegraphics[width=0.85\columnwidth]{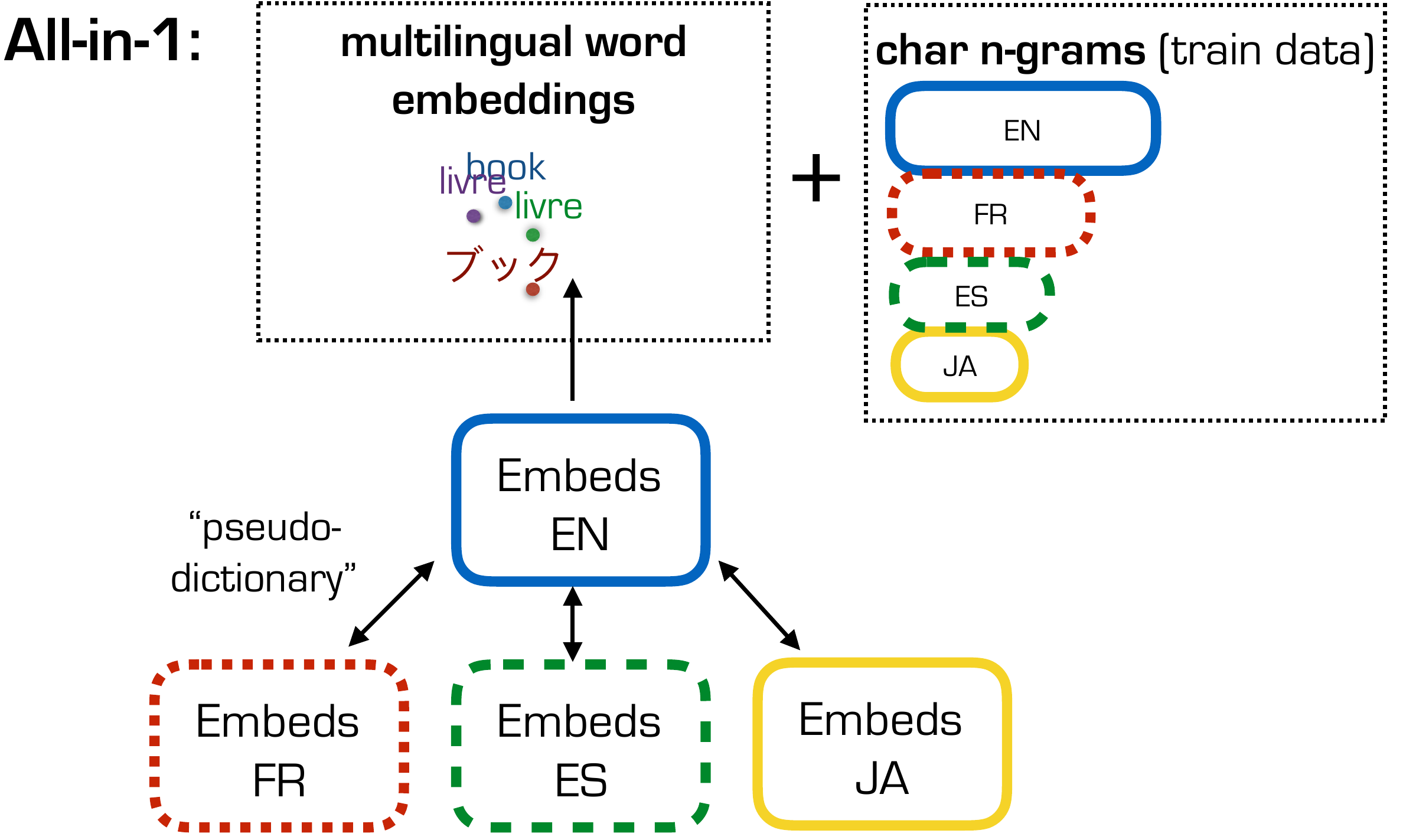}
\caption{Overview of our \textsc{All-In-1} model.}
\label{fig:model}
\end{figure}

Motivated by the goal to evaluate how good a single model for multiple languages fares, we decided to build a very simple model that can handle any of the four languages. We aimed at an approach that does \textit{not} require any language-specific processing (beyond tokenization) nor requires any parallel data. We set out to build a simple baseline, which turned out to be surprisingly effective. Our model is depicted in Figure~\ref{fig:model}.

Our key motivation is to provide a simple, general system as opposed to the usual ad-hoc setups one can expect in a multilingual shared task.  So we rely on character n-grams, word embeddings, and a traditional classifier, motivated as follows. 

First, character n-grams and traditional machine learning algorithms have proven successful for a variety of classification tasks, e.g., native language identification and language detection. In recent shared tasks simple traditional models outperformed deep neural approaches like CNNs or RNNs, e.g.,~\cite{medvedeva-kroon-plank:2017:VarDial,zampieri-EtAl:2017:VarDial,malmasi-EtAl:2017:BEA,kulmizev-EtAl:2017:BEA}. This motivated our choice of using a traditional model with character n-gram features.

Second, we build upon the recent success of multilingual embeddings. These are embedding spaces in which word types of different languages are embedded into the same high-dimensional space. Early approaches focus mainly on bilingual approaches, while recent research aims at mapping several languages into a single space. The body of literature is huge, but an excellent recent overview is given in~\newcite{xlingsurvey}. We chose a very simple and recently proposed method that does not rely on any parallel data~\cite{smith2017offline} and extend it to the multilingual case. In particular, the method falls under the broad umbrella of \textit{monolingual mappings}. These approaches first train monolingual embeddings on large unlabeled corpora for the single languages. They then learn linear mappings between the monolingual embeddings to map them to the same space. The approach we apply here is particularly interesting as it does not require parallel data (parallel sentences/documents or dictionaries) and is readily applicable to off-the-shelf embeddings. In brief, the approach aims at learning a transformation in which word vector spaces are orthogonal (by applying SVD) and it leverages so-called ``pseudo-dictionaries''. That is, the method first finds the common word types in two embedding spaces, and uses those as pivots to learn to align the two spaces (cf.\ further details in~\newcite{smith2017offline}). 

\section{Experimental Setup}
In this section we first describe the IJCNLP 2017 shared task 4\footnote{\url{https://sites.google.com/view/customer-feedback-analysis/}} including the data, the features, model and evaluation metrics. 

\subsection{Task Description}

The customer feedback analysis task~\cite{liuetal:2017:IJCNLP} is a short text classification task. Given a customer feedback message, the goal is to detect the type of customer feedback. For each message, the organizers provided one or more labels.  To give a more concrete idea of the data, the following are examples of the English dataset:
\begin{itemize}
\setlength\itemsep{1pt}
\item ``Still calls keep dropping with the new update'' (\textit{bug})
\item ``Room was grubby, mold on windows frames.'' (complaint)
\item ``The new update is amazing.'' (\textit{comment})
\item ``Needs more control s and tricks..'' (\textit{request})
\item ``Enjoy the sunshine!!'' (\textit{meaningless})
\end{itemize}

\subsection{Data}\label{sec:data}
The data stems from a joint ADAPT-Microsoft project. An overview of the provided dataset is given in Table~\ref{tbl:stats}. Notice that the available amount of data differs per language.  

\begin{table}\centering
\begin{tabular}{lrrrr}
\toprule
               & \textsc{En} & \textsc{Es} & \textsc{Fr} 
               & \textsc{Jp}\\
               \midrule
\textsc{Train} & 3066 & 1632 & 1951 & 1527\\
\textsc{Dev}  &   501 &  302 &  401 &  251\\
\textsc{Test} &   501 &  300 &  401 &  301 \\
\bottomrule
\end{tabular}
\caption{Overview of the dataset (instances).}
\label{tbl:stats}
\end{table}

\begin{figure}[h!]
\includegraphics[width=\columnwidth]{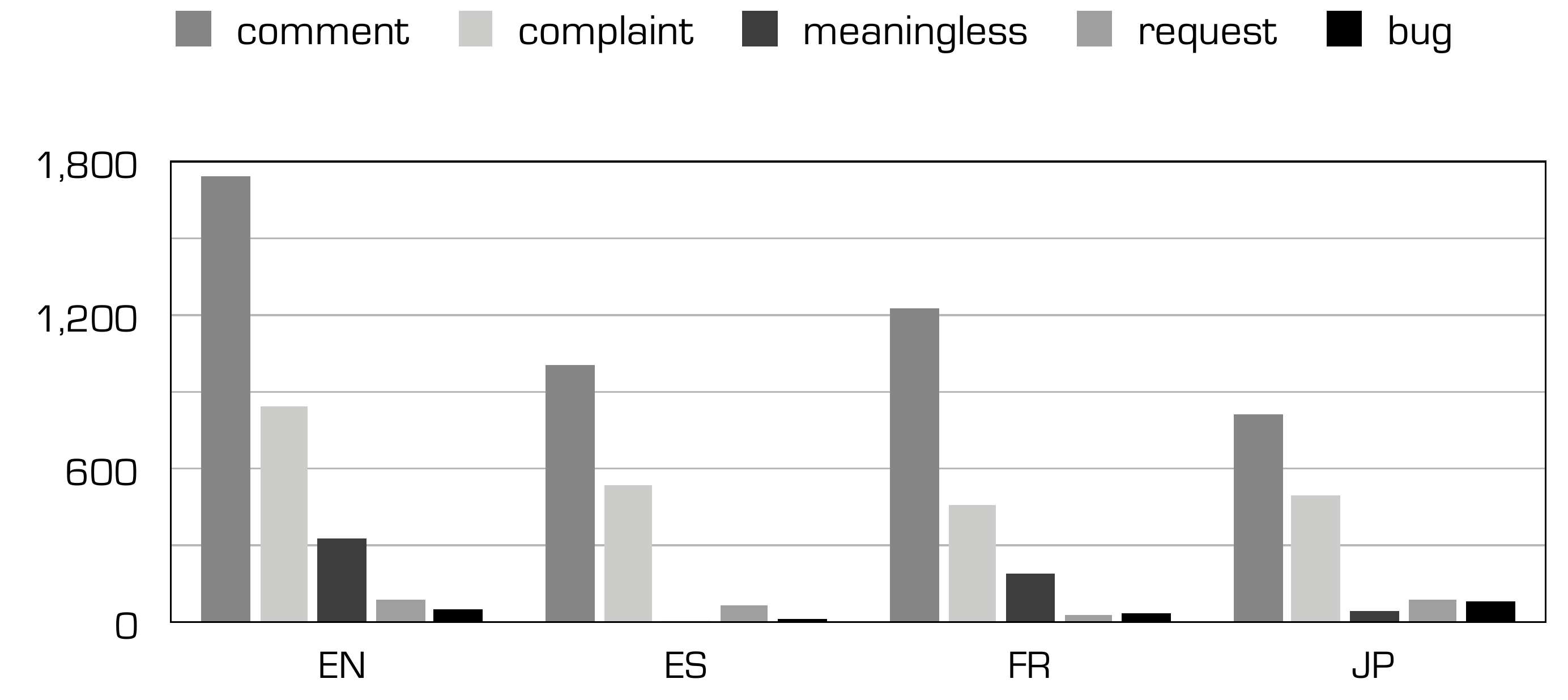}
\caption{Distribution of the labels per language.}
\label{fig:distrlabels}
\end{figure}

We treat the customer feedback analysis problem as a single-class classification task and actually ignore multi-label instances, as motivated next. The final label distribution for the data is given in Figure~\ref{fig:distrlabels}.

In initial investigations of the data we noticed that very few instances had multiple labels, e.g., ``\textit{comment},\textit{complaint}''. In the English training data this amounted to $\sim$4\% of the data. We decided to ignore those additional labels (just picked the first in case of multiple labels) and treat the problem as a single-class classification problem. This was motivated by the fact that some labels were expected to be easily confused. Finally, there were some labels in the data that did not map to any of the labels in the task description (i.e., `\textit{undetermined}', `\textit{undefined}',  `\textit{nonsense}' and `\textit{noneless}', they were presumably typos) so we mapped them all to the `\textit{meaningless}' label. This frames the task as a 5-class classification problem with the following classes:
\begin{itemize}
\setlength\itemsep{0pt}
\item \textit{bug}, 
\item \textit{comment}, 
\item \textit{complaint}, 
\item \textit{meaningless} and 
\item \textit{request}.
\end{itemize}

At test time the organizers additionally provided us with \textit{translations} of the three language-specific test datasets back to English. These translations were obtained by Google translate. This allowed us to evaluate our English model on the translations, to gauge whether translation is a viable alternative to training a multilingual model.

\subsection{Pre-processing}\label{sec:preproc}

We perform two simple preprocessing steps. First of all, we tokenize all data using off-the-shelf tokenizers. We use \texttt{tinysegmenter}\footnote{\url{https://pypi.python.org/pypi/tinysegmenter}} for Japanese and the NLTK \texttt{TweetTokenizer} for all other languages. The Japanese segmenter was crucial to get sufficient coverage from the word embeddings later. No additional preprocessing is performed.

\subsection{Multilingual Embeddings}
Word embeddings for single languages are readily available, for example the Polyglot\footnote{Despite their name the Polyglot embeddings are actually monolingual embeddings, but available for many languages.} or Facebook embeddings~\cite{bojanowski2016enriching}, which were recently released.

In this work we start from the monolingual embeddings provided by the Polyglot project~\cite{polyglot:2013}. We use the recently proposed approach based on SVD decomposition and a ``pseudo-dictionary''~\cite{smith2017offline} obtained from the monolingual embeddings to project embedding spaces. To extend their method from the bilingual to the multilingual case, we apply pair-wise projections by using English as pivot, similar in spirit to~\newcite{ammar2016massively}.  We took English as our development language. We also experimented with using larger embeddings (Facebook embeddings; larger in the sense of both trained on more data and having higher dimensionality), however, results were comparable while training time increased, therefore we decided to stick to the smaller 64-dimensional Polyglot embeddings.

\subsection{Model and Features} As classifier we use a traditional model, a Support Vector Machine (SVM) with linear kernel implemented in \texttt{scikit-learn}~\cite{scikit-learn}. We tune the regularization parameter $C$ on the English development set and keep the parameter fixed for the remaining experiments and all languages ($C=10$). 

We compared the SVM to \texttt{fastText}~\cite{joulin2016bag}. As we had expected \texttt{fastText} gave consistently lower performance, presumably because of the small amounts of training data. Therefore we did not further explore neural approaches.

Our features are character n-grams (3-10 grams, with binary tf-idf) and word embeddings. For the latter we use a simple continuous bag-of-word representation~\cite{collobert2011natural} based on averaging and min-max scaling.





Additionally, we experimented with adding Part-Of-Speech (POS) tags to our model. However, to keep in line with our goal to build a \textit{single system for all languages} we trained a single multilingual POS tagger by exploiting the projected multilingual embeddings. In particular, we trained a state-of-the-art bidirectional LSTM tagger~\cite{plank:ea:2016}\footnote{\url{https://github.com/bplank/bilstm-aux}} that uses both word and character representations on the concatenation of language-specific data provided from the Universal Dependencies data (version 1.2 for En, Fr and Es and version 2.0 data for Japanese, as the latter was not available in free-form in the earlier version). The word embeddings module of the tagger is initialized with the multilingual embeddings. We investigated POS n-grams (1 to 3 grams) as additional features.

\subsection{Evaluation} We decided to evaluate our model using weighted F1-score, i.e., the per-class F1 score is calculated and averaged by weighting each label by its support. Notice, since our setup deviates from the shared task setup (single-label versus multi-label classification), the final evaluation metric is  different. We will report on weighted F1-score for the development and test set (with simple macro averaging), but use Exact-Accuracy and Micro F1 over all labels when presenting official results on the test sets. The latter two metrics were part of the official evaluation metrics. For details we refer the reader to the shared task overview paper~\cite{liuetal:2017:IJCNLP}.

\section{Results}

We first present results on the provided development set, then on the official evaluation test set. 
\subsection{Results on Development}

\begin{table}
\resizebox{\columnwidth}{!}{

\begin{tabular}{lcccc|c}
\toprule
   & \textsc{En} & \textsc{Es}& \textsc{Fr}  & \textsc{Jp} & \textsc{Avg}\\
\midrule
\multicolumn{6}{c}{   \textsc{Monolingual Models} }\\
Embeds & 50.6 & 82.0 & 66.5 & 65.1 & 66.05\\
Words (W) & 66.1 & 86.9 & 73.2 & 73.6 & 74.95\\   
Chars (C) & 68.2 & 87.1 & 76.1 & 74.0 & 76.35\\   
W+Chars (C) & 65.9 & 87.7 & 75.7 & 74.0 & 75.82\\
C+Embeds$\ddagger$ & 66.1 & 86.6 & 76.5 & 77.1 & 76.58\\
W+C+Embeds & 65.9 & 87.8 & 75.6 & 76.8 & 76.52\\ 
\midrule
\multicolumn{6}{c}{\textsc{Bilingual Model}}\\
En+Es & 67.6 & 86.6 & -- & -- & --\\
En+Fr & 66.6 & -- & 77.8 & -- & --\\
En+Jp & 66.7 & -- & -- & 77.9 & --\\
\multicolumn{6}{c}{\textsc{Multilingual Model}}\\
En+Es+Fr & 68.3 & 87.0 & 77.9 & -- & --\\
\textsc{All-in-1}$\ddagger$ & 68.8 & 87.7 & 76.4 & 77.2 & 77.5\\

\midrule
\textsc{All-in-1}+POS & 68.4 & 86.0 & 74.4 & 74.5 & 75.8\\
\bottomrule
\end{tabular}
}
\caption{Results on the development data, weighted F1. \textsc{Monolingual}: per-language model; \textsc{Multilingual}: \textsc{All-In-1} (with C+Embeds features trained on En+Es+Fr+Jp). $\ddagger$ indicates submitted systems.}
\label{tbl:resultsdev}
\end{table}

First of all, we evaluated different feature representations. As shown in Table~\ref{tbl:resultsdev} character n-grams alone prove very effective, outperforming word n-grams and word embeddings alone. Overall simple character n-grams (C) in isolation are often more beneficial than word and character n-grams together, albeit for some languages results are close. The best representation are character n-grams with word embeddings. This representation provides the basis for our multilingual model which relies on multilingual embeddings. The two officially submitted models both use character n-grams (3-10) and word embeddings. Our first official submission, \textsc{Monolingual} is the per-language trained model using this representation.

Next we investigated adding more languages to the model, by relying on the multilingual embeddings as bridge. For instance in Table~\ref{tbl:resultsdev}, the model indicated as En+Es is a character and word embedding-based SVM trained using bilingual embeddings created by mapping the two monolingual embeddings onto the same space and using both the English and Spanish training material. As the results show, using multiple languages can improve over the in-language development performance of the character+embedding model. However, the bilingual models are still only able to handle pairs of languages. We therefore mapped all embeddings to a common space and train a single multilingual \textsc{All-in-1} model on the union of all training data. This is the second model that we submitted to the shared task. As we can see from the development data, on average the multilingual model shows promising, overall (macro average) outperforming the single language-specific models. However, the multilingual model does not consistently fare better than single models, for example on French a monolingual model would be more beneficial. 

Adding POS tags did not help (cf.\ Table~\ref{tbl:resultsdev}), actually dropped performance. We disregard this feature for the final official runs.

\subsection{Test Performance}

\begin{table}\resizebox{\columnwidth}{!}{
\begin{tabular}{lcccc|c}
\toprule
   & \textsc{En} & \textsc{Es}& \textsc{Fr}  & \textsc{Jp} & \textsc{Avg}\\
\midrule
\textsc{Monoling}  & \textbf{68.6} & 88.2 & \textbf{76.1} & 74.3 & 76.8\\
\textsc{Multiling}       & 68.1 & \textbf{88.7} & 73.9 & \textbf{76.7} & \textbf{76.9}\\
\textsc{Translate} & -- & 83.4 & 69.5 & 61.6 & --\\
\bottomrule
\end{tabular}
}
\caption{Results on the test data, weighted F1. \textsc{Monoling}:  monolingual models. \textsc{Multiling}: the multilingual \textsc{All-In-1} model. \textsc{Trans}: translated targets to English and classified with \textsc{En} model.}
\label{tbl:resultstestwF1}
\end{table}

We trained the final models on the concatenation of \textsc{Train} and \textsc{Dev} data. The results on the test set (using our internally used weighted F1 metric) are given in Table~\ref{tbl:resultstestwF1}. 

There are two take-away points from the main results: First, we see a positive transfer for languages with little data, i.e., the single multilingual model outperforms the language-specific models on the two languages (Spanish and Japanese) which have the least amount of training data. Overall results between the monolingual and multilingual model are close, but the advantage of our multilingual \textsc{All-in-1} approach is that it is a single model that can be applied to all four languages. Second, automatic translation harms, the performance of the \textsc{EN} model on the translated data is substantially lower than the respective in-language model. We could investigate this as the organizers provided us with translations of French, Spanish and Japanese back to English.

\begin{table}[h!]
\resizebox{\columnwidth}{!}{

\begin{tabular}{lcccccc}
\toprule
      & \textsc{En} & \textsc{Es} & \textsc{Fr} & \textsc{Jp} & \textsc{Avg} \\
\midrule
Ours (\textsc{Multiling} ) & 68.60 & \textbf{88.63} & 71.50 & \textbf{75.00}& \textbf{76.04}\\
Ours (\textsc{Monoling}) & 68.80 & 88.29 & \textbf{73.75} & 73.33 & 75.93\\
YNU-HPP-glove$\dagger$ & \textbf{71.00} & --& --& -- & --\\
FYZU-bilstmcnn & 70.80 & -- & -- & -- & -- \\  
IITP-CNN/RNN & 70.00 & 85.62 & 69.00 & 63.00 & 71.90\\
TJ-single-CNN$\dagger$ & 67.40 & -- & -- & --\\
Baseline & 48.80 & 77.26 & 54.75 & 56.67 & 59.37\\
\bottomrule
\end{tabular}
}
\caption{Final test set results (Exact accuracy) for top 5 teams (ranked by macro average accuracy). Rankings for micro F1 are similar, we refer to the shared task paper for details. Winning system per language in bold. $\dagger$: no system description available at the time of writing this description paper.}
\label{tbl:resultstop}
\end{table}

Averaged over all languages our system ranked first, cf.\ Table~\ref{tbl:resultstop} for the results of the top 5 submissions. The multilingual model reaches the overall best exact accuracy, for two languages training a in-language model would be slightly more beneficial at the cost of maintaining a separate model.  The similarity-based baseline provided by the organizers\footnote{``Using n-grams (n=1,2,3) to compute sentence similarity (which is normalized by the length of sentence). Use the tag(s) of the most similar sentence in training set as predicted tag(s) of a sentence in the test set.''} is considerably lower. 

Our system was outperformed on English by three teams, most of which focused only on English. Unfortunately at the time of writing there is no system description available for most other top systems, so that we cannot say whether they used more English-specific features. From the system names of other teams we may infer that most teams used neural approaches, and they score worse than our SVM-based system.

The per-label breakdown of our systems on the official test data (using micro F1 as calculated by the organizers) is given in Table~\ref{tbl:resultstest}. Unsurprisingly less frequent labels are more difficult to predict. 

\begin{table}
\resizebox{\columnwidth}{!}{

\begin{tabular}{lrrrrr}
\toprule
      & \textit{comm} & \textit{compl} & \textit{req} & \textit{ml} & \textit{bug} \\
\midrule
\textsc{En} (\textsc{Monoling})    & \textbf{82.3} & 64.4 & \textbf{60.0} & 27.5 & 0 \\
\textsc{En} (\textsc{Multiling})   & 82.0 & \textbf{65.0} & 42.1 & \textbf{28.6} & 0 \\
\midrule
\textsc{Es} (\textsc{Monoling})    & 93.3 & 75.2 & \textbf{72.7} & 0 & 0 \\
\textsc{Es} (\textsc{Multiling})   & \textbf{93.5} & \textbf{76.2} & 66.6 & 0 & \textbf{66.6} \\
\textsc{Es} (\textsc{Translate})   & 92.6 & 67.2 & 11.8 & 0 & 0\\
\midrule
\textsc{Fr} (\textsc{Monoling})    & \textbf{86.4} & \textbf{65.6 }& 14.3 & \textbf{47.6} & \textbf{54.5}  \\
\textsc{Fr} (\textsc{Multiling})   & 85.5 & 61.5 & \textbf{16.6} & 41.2 & 50.0 \\
\textsc{Fr} (\textsc{Translate})   & 82.9 & 58.9 & 16.6 & 34.5 & 0\\
\midrule
\textsc{Jp} (\textsc{Monoling})   & 85.7 & \textbf{67.8} & 55.8 & 0 & \textbf{50.0} \\
\textsc{Jp} (\textsc{Multiling} )  & \textbf{87.0} & \textbf{67.8} & \textbf{65.2} & 0 & \textbf{50.0} \\
\textsc{Jp} (\textsc{Translate} )  & 76.5 & 61.3 & 7.2 & 0 & 0 \\
\bottomrule
\end{tabular}
}
\caption{Test set results (F1) per category (\textit{comment (comm)}, \textit{complaint (compl)}, \textit{request (req)}, \textit{meaningless (ml)} and \textit{bug}), official evaluation.}
\label{tbl:resultstest}
\end{table}

\section{Conclusions}

We presented a simple model that can effectively handle multiple languages in a single system. The model is based on a traditional SVM, character n-grams and multilingual embeddings. The model ranked first in the shared task of customer feedback analysis, outperforming other approaches that mostly relied on deep neural networks. 

There are two take-away messages of this work: 1) multilingual embeddings are very promising\footnote{Our study is limited to using a single multilingual embedding method and craves for evaluating alternatives!} to build single multilingual models; and 2) it is important to compare deep learning methods to simple traditional baselines; while deep approaches are undoubtedly very attractive (and fun!), we always deem it important to compare deep neural to traditional approaches, as the latter often turn out to be surprisingly effective. Doing so will add to the literature and help to shed more light on understanding why and when this is the case.

\section*{Acknowledgments}
I would like to thank the organizers, in particular Chao-Hong Liu, for his quick replies. I also thank Rob van der Goot, H\'{e}ctor Mart\'{i}nez Alonso and Malvina Nissim for valuable comments on earlier drafts of this paper. 

\bibliography{biblio}
\bibliographystyle{ijcnlp2017}

\end{document}